\documentclass{trbunofficial}
\usepackage{latexsym}
\usepackage{graphicx}
\usepackage{amsfonts,amssymb,amsmath}
\usepackage{hyperref}

\usepackage{graphics} 
\usepackage{dblfloatfix} 
\usepackage[ruled,vlined]{algorithm2e}
\usepackage[utf8]{inputenc} 
\usepackage[T1]{fontenc}    
\usepackage{hyperref}       
\usepackage{url}          
\usepackage{booktabs}      
\usepackage{nicefrac}       
\usepackage{microtype}     
\usepackage{caption}
\usepackage{subcaption}
\usepackage{catchfile}
\usepackage{totcount}
\usepackage{multirow}
\usepackage{array}

\newcolumntype{P}{>{\centering\arraybackslash}m{1.45cm}}
\newcolumntype{Q}{>{\centering\arraybackslash}m{0.90cm}}

\nolinenumbers
\title{An Assessment of Safety-Based Driver Behavior Modeling in Microscopic Simulation Utilizing Real-Time Vehicle Trajectories}

\author{\textbf{Awad Abdelhalim}, Corresponding Author  \\
  Postdoctoral Research Associate \\
  Department of Urban Studies and Planning \\
  Massachusetts Institute of Technology, Cambridge, MA 02139 \\
  Email: awadt@mit.edu \\
  \hfill\break
  \textbf{Montasir Abbas}\\
  Professor \\
  Virginia Polytechnic Institute and State University \\
  301 Patton Hall, Virginia Tech, Blacksburg, VA 24061 \\
  Tel: 540-231-9002 \\
  Email: abbas@vt.edu
}

\begin{document}
\maketitle

\section{Abstract}

Accurate representation of observed driving behavior is critical for effectively evaluating safety and performance interventions in simulation modeling. In this study, we implement and evaluate a safety-based Optimal Velocity Model (OVM) to provide a high-fidelity replication of safety-critical behavior in microscopic simulation and showcase its implications for safety-focused assessments of traffic control strategies. A comprehensive simulation model is created for the site of study in PTV VISSIM utilizing detailed vehicle trajectory information extracted from real-time video inference, which are also used to calibrate the parameters of the safety-based OVM to replicate the observed driving behavior in the site of study. The calibrated model is then incorporated as an external driver model that overtakes VISSIM's default Wiedemann 74 model during simulated car-following episodes. The results of the preliminary analysis show the significant improvements achieved by using our model in replicating the existing safety conflicts observed at the site of the study. We then utilize this improved representation of the status quo to assess the potential impact of different scenarios of signal control and speed limit enforcement in reducing those existing conflicts by up to 23\%. The results of this study showcase the considerable improvements that can be achieved by utilizing data-driven car-following behavior modeling, and the workflow presented provides an end-to-end, scalable, automated, and generalizable approach for replicating the existing driving behavior observed at a site of interest in microscopic simulation by utilizing vehicle trajectories efficiently extracted via roadside video inference.

\hfill\break%
\noindent\textit{Keywords}: Driver behavior modeling, optimal velocity model, microscopic simulation.
\newpage

\section{INTRODUCTION}
Traffic modeling and simulation is one of the most powerful tools at the disposal of transportation engineers today. The modeling and simulation of a specific site, corridor, or network of interest provides transportation practitioners with an efficient and effective method to assess the current performance and evaluate any proposed performance or safety interventions. Simulation modeling can be sub-categorized into three types:
\begin{itemize}
    \item Macroscopic Simulation, which provides a high-level representation of traffic streams, 
    \item Microscopic Simulation, where vehicles are modeled at an individual level, and
    \item Mesoscopic Simulation, which aims to utilize a combination of macro and microscopic simulation methods.
\end{itemize}

While macroscopic and mesoscopic are the more common choices of modeling when simulating larger networks, microscopic simulation is more often used for studying the traffic flow in smaller areas, in greater detail. In microscopic simulation, however, the individual vehicles' behavior is based on mathematically-derived driver behavior models that require a tedious process of data collection and parameter calibration to better replicate the observed behavior in the specific area of interest. This has generally been the challenge for the wide-scale adoption of microscopic simulation.

PTV's Verkehr In Städten SIMulationsmodel (VISSIM) is a discrete, stochastic, time step-based simulation software that is one of the most popular tools used for traffic simulation due to its extensive capabilities for modeling and evaluating different components of the transportation ecosystem (vehicles, public transit, pedestrians, cyclists, roadside infrastructure, etc.). VISSIM's base car-following model is based on the Wiedemann 74 and 99 psycho-physical models \cite{wiedemann1991modelling}. The car-following behavior of vehicles in a VISSIM simulation is modified through ten driver behavior parameters (labeled CC0 – CC9) that represent different thresholds of four assumed driving states: free-driving, approaching, following, and braking. For each of those modes, the desired instantaneous acceleration is a result of the vehicle's current speed, the speed and distance differences between the vehicle and the lead vehicle, and specific characteristics of the driver and the vehicle. The challenge with the Wiedemann model (and other conventional driver behavior models, e.g. Gazis-Herman-Rothery, Fritzsche) is that they assume that drivers make longitudinal decisions (i.e. acceleration) based on assumed momentary stimuli inputs. These models do not account for the driver’s planned decision process, where a driver could perceive an impending danger, and plans to avoid that danger, then executes that plan in multiple time steps. This is especially important when modeling safety-critical or close-to-critical situations, when an ego vehicle approaches a vehicle that slows down, forcing the driver of the ego vehicle to start a process of avoidance or adjustment in their speed trajectory. The failure to capture this behavior leads to an inaccurate representation of traffic when modeled in simulation, hindering the ability to evaluate safety interventions.

\subsection{Objective and Contributions}
This study aims to showcase the improvements that can be achieved in VISSIM simulation of safety-critical behavior utilizing real-time, safety-based driver behavior modeling. The real-time safety-based Optimal Velocity Model, which was proposed in a recent study \cite{abdelhalim2021safety} is calibrated for the site of study using vehicle trajectories extracted from a roadside camera and implemented in VISSIM as an external driver model. We compare the performance of our model to the default VISSIM model in terms of replicating the existing safety-critical behavior at the area of study. We then assess the impact of different scenarios of signal control and speed enforcement in mitigating the simulated safety concerns. The following sections of this paper are as follows: (a) a survey of related literature on driver behavior modeling in microscopic simulation and trajectory-based traffic safety, (b) a detailed breakdown of the VISSIM model development and external driver model implementation, (c) results and analyses of the case study, and, (d) discussion and conclusions.

\subsection{Related Work}

\subsubsection{Driver Behavior and Safety Modeling in Microscopic Simulation}
Over the past two decades, the growing adoption of microscopic simulation software has provided researchers with powerful means for evaluating different components of transportation networks and their interactions. The simulated vehicle interactions, which form the basis of microscopic driving behavior models, allowed for the assessment of traffic safety and evaluation of countermeasures as opposed to solely relying on analyzing crash data from police reports. The Surrogate Safety Assessment Model (SSAM) \cite{gettman2008surrogate} is the prime example of such efforts. Using vehicle trajectories exported from microsimulation as an input, SSAM's approach utilizes conflict analysis for near misses, in terms of time-to-collision (TTC) and post encroachment time (PET), to infer surrogate measures of actual crash data in terms of frequency, types, and severity of crashes. SSAM, however, is dependent on the outputs generated by microsimulation. Hence, adequate parameter calibration for the underlying driver behavior model used to generate the vehicle trajectories is a crucial step that the lack thereof was found to result in a significant deviation of simulated safety conflicts from the observed behavior \cite{vasconcelos2014validation}, \cite{huang2013identifying}, \cite{dijkstra2010calculated}.

Given the increasing popularity of VISSIM over the years, numerous studies have been conducted to assess the sensitivity of calibrating its underlying Wiedemann car-following model to better mimic the existing traffic flow and behavior and result in more realistic simulated safety conflicts. Fellendorf and Vortisch \cite{fellendorf2001validation} conducted one of the earliest studies, where they assessed the impact of the driver model calibration on both a micro and macroscopic level. The results of their study indicated that a well-calibrated model can replicate traffic flow with high accuracy, with the caveat that driver behavior models should at least take national and regional regulations and driving styles into account to produce reliable representations in simulation. Lownes and Machemehl \cite{lownes2006vissim, lownes2006sensitivity} conducted a sensitivity analysis to assess the impact of modifying VISSIM's driver behavior parameters on simulated traffic flow capacity. Their work has highlighted the impact of changes on driver behavior models at a microscopic level on traffic flow at a macroscopic level, further making the case for the necessity of parameter calibration to ensure adequate representation of traffic modeled in the simulation. Other studies have also highlighted the necessity of taking into account the varying traffic patterns to be modeled and the specific considerations needed for calibrating VISSIM's driver behavior parameters accordingly, such cases include modeling traffic at urban intersections \cite{manjunatha2013methodology}, \cite{arafat2020data} and freeway merge areas \cite{fan2013using}. The work of Fan et. al. \cite{fan2013using} particularly stands out in this category, as VISSIM's default driver behavior parameters were calibrated for freeway driving, yet separate parameter calibration for merging sections specifically was necessary, with the results highlighting the significant improvements obtained. This was attributed to the default VISSIM model parameters being calibrated from driving data from the German autobahn, while the study was conducted in China, hence the driving behavior parameters were not entirely transferable.

The transferability of the calibrated VISSIM driver behavior model parameters was assessed by Essa and Sayed \cite{essa2015transferability}, who calibrated the VISSIM model parameters to maximize the correlation between observed and simulated traffic conflicts in one intersection and tested the calibrated model on a nearby intersection. Their study concluded that while not all of the calibrated model parameters are fully transferable, the calibrated model still significantly outperforms the default model in terms of providing a better correlation between the observed and simulated traffic conflicts. Hence, any safety assessment without proper model calibration should be avoided. Essa and Tarek further highlighted this need for rigorous calibration and the limitations of simulation-based safety assessment in other studies \cite{essa2015simulated}, \cite{essa2020comparison}.

The extensive literature in VISSIM, and microscopic model parameter calibration in general, lacks a clear and transferable framework to overcome the shortcomings of site-by-site data collection and calibration of driver model parameters to reduce errors versus observed data. In recent years, researchers have opted to utilize neural network models to overpass this tedious process \cite{otkovic2020validation}, \cite{naing2021data}. And while such methods succeed in producing improved results, they lack interpretability, unlike the traditional driving behavior models which are based on traffic flow theory fundamentals. There exists, therefore, a need for a generalizable driving behavior modeling approach for simulation to better mimic the vehicle trajectories being simulated and their resulting safety conflicts, allowing for a more accurate assessment of proposed safety interventions. To overcome this gap identified in driver behavior modeling calibration and its application in VISSIM, we propose an implementation of a data-driven high-fidelity external driver behavior model in VISSIM and showcase its utilization in a case study to assess appropriate safety interventions. The proposed external driver model is calibrated from video inference for the site of interest, providing an end-to-end scalable, automated, and generalizable workflow for replicating the observed behavior in microscopic simulation.

\section{METHODOLOGY}
\subsection{Area of Study, Data Collection, and the VISSIM Model}
\begin{figure}[!h]
\begin{subfigure}{.48\textwidth}
  \centering
  \includegraphics[width= \textwidth, height= 2.0 in]{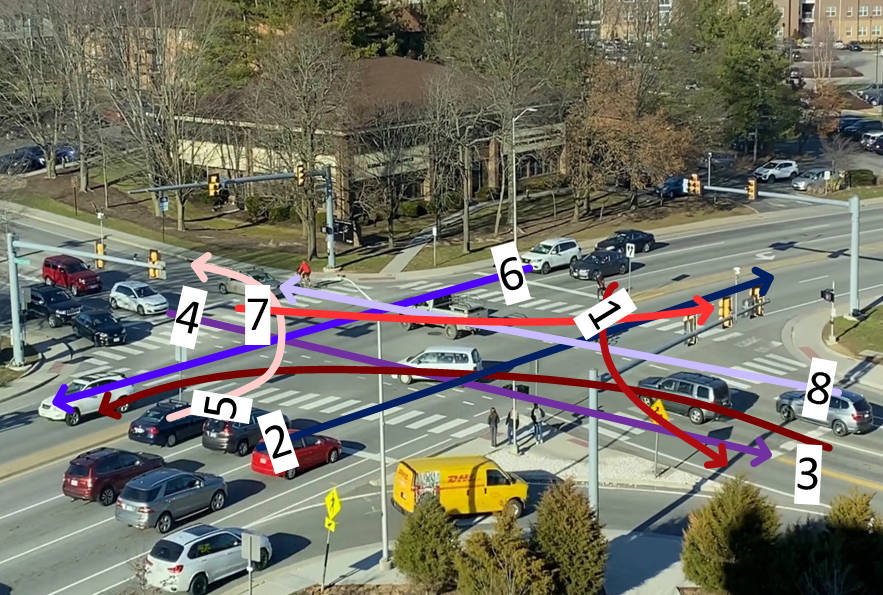}
  \caption{\label{fig:sub2} Site of study and NEMA movement labels.}
\end{subfigure}
\begin{subfigure}{.48\textwidth}
  \centering
  \includegraphics[width= \textwidth, height= 2.0 in]{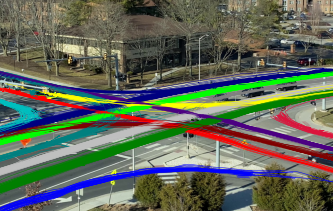}
  \caption{\label{fig:sub1} Trajectories and movement classification.}
\end{subfigure}
\begin{subfigure}{\textwidth}
  \centering
  \includegraphics[width= 0.50\textwidth, height= 2.0 in]{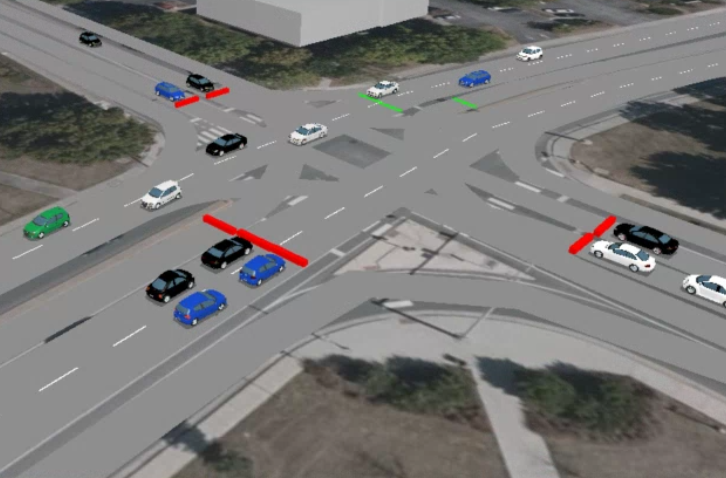}
  \caption{\label{fig:sub1} The VISSIM model developed for this study.}
\end{subfigure}
\caption{\label{fig:area} Site of study and tracked vehicle trajectories.}
\end{figure}

\begin{figure}[!h]
\centering
\includegraphics[scale= 0.48]{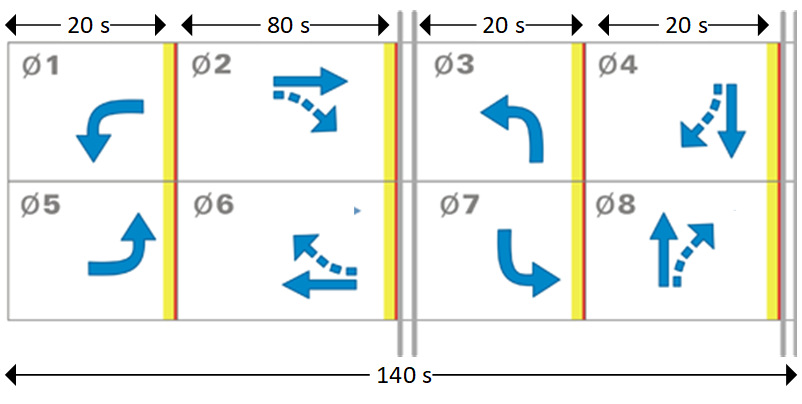}
\caption{\label{fig:nema_diagram} NEMA movements and on-site signal timing plan.}
\end{figure}

In a recent study \cite{abdelhalim2021safety}, we utilized an extended VT-Lane framework to obtain the trajectories and calibrate a data-driven safety-based driver behavior model for the area of study. Previous works detail the implementation and evaluation of the computer-vision-based trajectory tracking framework \cite{abdelhalim2020vt, abdelhalim2020towards, abdelhalim2021framework}. Figure \ref{fig:area} shows the area of this study from the perspective of the roadside camera used to obtain the video data, illustrating the NEMA movement enumeration for the site. The traffic volumes and speed distributions obtained from the inference of 1-hour PM traffic footage were modeled in VISSIM alongside the existing on-site signal timing plan illustrated in Figure \ref{fig:nema_diagram}.

\subsection{Incorporating the Safety-Based Driver Model}
\begin{figure}[!h]
\centering
\includegraphics[width= 0.51\textwidth]{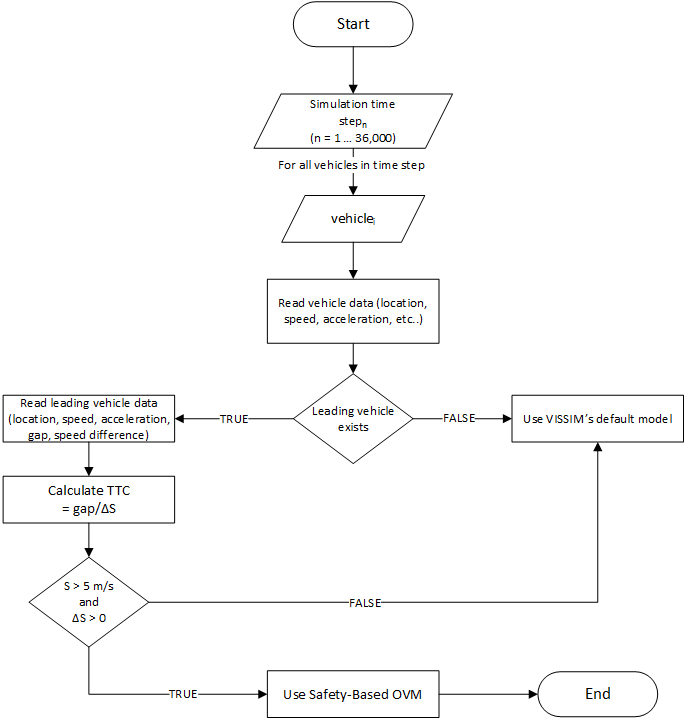}
\caption{\label{fig:flowchart} VISSIM driver behavior model selection logic.}
\end{figure}

VISSIM's external driver model module is utilized to incorporate our calibrated safety-based OVM. The C++ generated Dynamic Link Library (DLL) file is executed during VISSIM's simulation runs once every simulation step for all vehicles present in the network. All vehicles in the network are controlled by the default Wiedemann model that VISSIM utilizes unless they are engaged in car-following episodes. In this study, a vehicle is considered in car-following if it is moving at a speed $\geq$ 5 m/sec (18 km/hr), and there exists a slower leading vehicle moving in the same lane. The threshold ensures that the model is not utilized for vehicles that are stopping at the intersection due to signal control, which could be assumed to be in a car-following model based on estimated TTC. Figure \ref{fig:flowchart} shows a flowchart with the code execution logic for driver behavior model selection in VISSIM, which is detailed in Algorithm \ref{algorithm:external}.

\newpage
The widely used optimal velocity model first proposed by Bando et al. \cite{bando1995dynamical} calculates the speed of the following vehicle based on the distance gap given by Equation \ref{eq:ovm_speed_orig}, from which the desired acceleration is calculated using Equation \ref{eq:ovm_acc}.

\vspace{-5pt}
\begin{equation}
  v_{opt}(s) = v_o \frac{\text{tanh}\left(\frac{s}{\Delta s} -\beta\right) + \text{tanh} \ \beta}{1+ \text{tanh} \ \beta}
  \label{eq:ovm_speed_orig}
\end{equation}

\begin{equation}
  \dot{v} = \frac{v_{opt}(s) - v}{\tau}
  \label{eq:ovm_acc}
\end{equation}

\noindent Where: \\
$s$ = Distance gap between vehicles in a car-following episode ($m$).\\
$v_{opt}(s)$ = The theoretical optimal velocity for a given distance gap ($km/hr$).\\
$v_o$ = Desired speed ($km/hr$), $\Delta s$ = Transition width ($m$).\\
$\beta$ = Form Factor, $\dot{v}$ = OVM acceleration ($km/hr/sec$).\\
$v$ = Actual speed of a following vehicle ($km/hr$).\\
$\tau$ = Adaptation time ($sec$).

For our modified safety-based OVM \cite{abdelhalim2021safety} which is incorporated in VISSIM as an external driver model for this study, the instantaneous optimal velocity and resulting desired acceleration based on real-time TTC are calculated using the following equations:

\vspace{-10pt}
\begin{equation}
  v_{opt}(ttc)_{i,n} = v_o \frac{\text{tanh}\left(\frac{ttc_{i,n}}{\Delta s} -\beta\right) + \text{tanh} \ \beta}{1+ \text{tanh} \ \beta}
  \label{eq:ovm_speed}
\end{equation}

\vspace{-20pt}
\begin{equation}
  \dot{v} = (1-\alpha)\frac{v_{opt_{i,n}} - v_{i,n}}{\tau} + \alpha f(ttc_{observed})
  \label{eq:ovm_mod}
\end{equation}

\noindent Where: \\
$ttc_{i,n}$ = Instantaneous time-to-collision between vehicle$_i$ and preceding vehicle in a car-following episode at simulation time step $n$ ($sec$).\\ 
$v_{opt}(ttc)_{i,n}$ = The theoretical optimal velocity for vehicle $i$ during simulation time step $n$ ($km/hr$).\\
$v_o$ = Desired speed ($km/hr$), $\Delta s$ = Transition width ($sec$).\\
$\beta$ = Form Factor, $\dot{v_{i,n}}$ = Desired acceleration for vehicle $i$ during time step $n$ ($km/hr/sec$).\\
$v_{i,n}$ = Actual speed of vehicle $i$ during simulation time step $n$ ($km/hr$).\\
$\tau$ = Adaptation time ($simulation \ time \ steps$).

The weighted observed acceleration, $\alpha f(ttc_{observed})$, allows our OVM to implicitly learn location-specific driving behavior characteristics from a subset of the observed driving behavior on-site to supplement the base OVM model's assumptions while maintaining the benefits of interpretability of the remaining model parameters (e.g., the calibrated model had a lower desired speed for vehicles executing turning movements compared to through moving vehicles, which is logical). We utilized VT-Lane's ability to classify the vehicles' movements across the intersection to calibrate separate model parameters for through-moving vehicles and vehicles executing turning movements at the intersection. The parameter calibration process for the model and substantial improvements achieved compared to the base OVM are detailed in \cite{abdelhalim2021safety}. The calibrated model parameters for each movement type utilized by the external driver model in this study are detailed in Table \ref{tab:params}. Algorithm \ref{algorithm:external} details the steps for the external driver model application during each simulation time step, where it is applied to all the vehicles in simulation.

\begin{table}[!h]
  \renewcommand{\arraystretch}{0.75}
  \centering
  \caption{Optimal Parameters For The Calibrated Safety-Based OVM}
    \begin{tabular}{lcc}
    \toprule
    \multirow{2}{*}{} & \multicolumn{2}{c}{\textbf{Vehicle Movement}} \\
\cmidrule{2-3} & \textbf{Through} & \textbf{Turning} \\
    \midrule
    \textbf{$\Delta s$} & 4.63 & 4.76 \\
    \midrule
    \textbf{$\beta$} & 1.10 & 0.10\\
    \midrule
    \textbf{$\tau$} & 6.11 & 5.83\\
    \midrule
    \textbf{$v_o$} & 38.20 & 31.95\\
    \midrule
    \textbf{$\alpha$} & 0.18 & 0.01\\
    \bottomrule
    \end{tabular}%
  \label{tab:params}%
\end{table}%

\begin{algorithm}
\SetAlgoLined
\KwResult{Desired acceleration of vehicles in VISSIM simulation.}
 initialization\;
 \For{$time\_step_n$}{
    \For{$vehicle_i$}{
        \eIf{$speed_{i,n}\ < \ 5 \ (m/s)$}{
            $acceleration_{i,n} \leftarrow{} Default$\;
            }{
            \eIf{$vehicle_{i,n}\ \exists\ \Delta\ Distance\ \And\ \iff\ \Delta\ Speed \in \mathbb{R}^+$}{
            $v_{opt_{i,n}} \leftarrow{} v_o \frac{\text{tanh}\left(\frac{ttc_{i,n}}{\Delta s} -\beta\right) \ + \ \text{tanh} \ \beta}{1+ \text{tanh} \ \beta}$\;
            $acceleration_{i,n} \leftarrow{} {\scriptsize (1-\alpha)}\frac{v_{opt_{i,n}} - \ speed_{i,n}}{\tau} + \alpha f(ttc_{observed})$\;

                }
            {
            $acceleration_{i,n} \leftarrow{} Default$\;
            }
        }}}
\caption{VISSIM External Driver Behavior Model}
\label{algorithm:external}
\end{algorithm}

\section{RESULTS AND ANALYSIS}
\subsection{Observed Vehicle Behavior and Safety Conflicts}
\begin{figure}[!h]
\begin{subfigure}{.48\textwidth}
  \centering
  \includegraphics[width= \textwidth, height= 2.4 in]{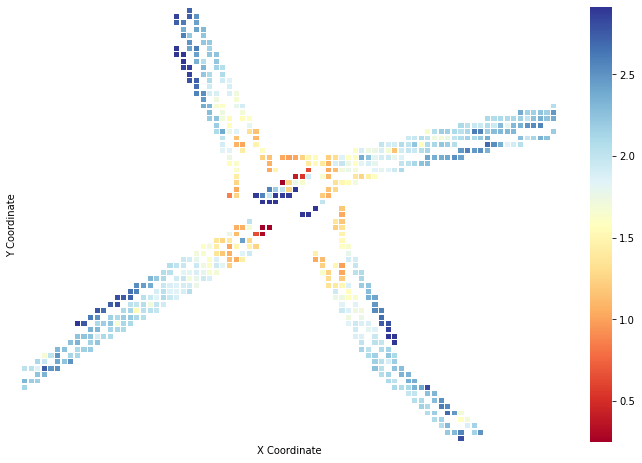}
  \caption{\label{fig:sub2} VISSIM's base Wiedemann model.}
\end{subfigure}
\begin{subfigure}{.48\textwidth}
  \centering
  \includegraphics[width= \textwidth, height= 2.4 in]{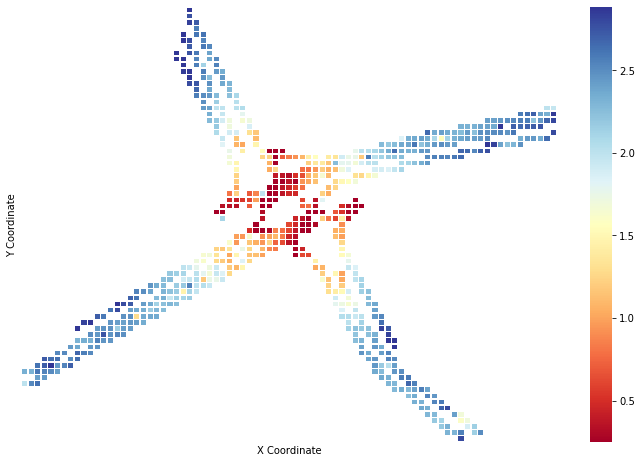}
  \caption{\label{fig:sub1} Safety-Based Optimal Velocity Model.}
\end{subfigure}
\caption{\label{fig:ttc_critical} Observed average TTC (where instantaneous TTC $\leq$ 3s) for simulation model.}
\end{figure}

A 1-hour simulation (preceded by a 15-minute warm-up) was conducted in VISSIM and the instantaneous time-to-collision for vehicles involved in car-following episodes was calculated and stored in an external database. Figure \ref{fig:ttc_critical} illustrates the location of safety conflicts and the associated average TTC for car-following instances where the TTC was $\leq$ 3 seconds for each location. X and Y coordinates are respectively the local longitude/latitude in the VISSIM model.


It can be clearly observed that VISSIM's base Wiedemann model is extremely conservative in generating safety conflicts during car-following episodes. Even when instances of low TTCs do occur, the vast majority take place as vehicles approach the stop bars at the intersections' four approaches. The base VISSIM model generates minimal safety conflicts as vehicles traverse the intersection and as they continue to accelerate downstream, where following vehicles seem to always be accelerating at a rate slower than the leading vehicles, leading to a severe underestimation of safety conflicts as reported by numerous studies in the literature. Our Safety-Based OVM that was incorporated in VISSIM generates a more realistic driving behavior that would be expected at an urban intersection, especially as vehicles accelerate to clear the intersection and at the merge areas of through lanes with turning lanes as clearly illustrated in Figure \ref{fig:ttc_critical} (b).

This distinction between VISSIM's default behavior and the external driver model is most evident when looking at safety-critical situations where the TTC between vehicles involved in a car-following episode momentarily drops to 1 second or less as can be seen in the figure. Out of 2667 vehicles in the 1-hour simulation, 31 vehicles (1.1\%) had safety-critical car-following instances using VISSIM's default model, compared to 210 (7.8\%) when utilizing the external driver model, where an instance is one simulation timestep (0.10 seconds).

\subsection{Impact Assessment of Mitigation Measures}
Based on this replication of observed driving behavior, we assess the impact of different mitigation measures on reducing the number of simulated conflicts. The area of study has a posted speed limit of 35 mph (55 km/hr). The observed speeds from video inference that were utilized in creating this simulation model had an 80th percentile speed of 35 mph (55 km/hr) and 95th percentile speed of 47 mph (75 km/hr), which is expected as there is no speed enforcement on-site and drivers typically drive above this posted speed limit. We assess the impact of enforcing speeds at 65 km/hr and 55 km/hr, respectively. Those enforcement measures are assessed with three scenarios of signal control, the current signal control previously illustrated in Figure \ref{fig:nema_diagram}, a scenario with a half-cycle length (70 seconds instead of 140), and a split-phasing signal control scenario. A sensitivity analysis for those enforcement and signal control scenarios is conducted for the observed, as well as half, and double the traffic volumes. For brevity, the conventions for the different scenarios used hereafter are shown in Table \ref{tab:scenarios}. Signal control scenarios are color-coded and labeled by name.

\begin{table}[!h]
  \renewcommand{\arraystretch}{1.10}
  \centering
  \caption{Conventions for Volume/Speed States}
    \begin{tabular}{|l|c|c|c|}
    \hline
    \textbf{$\downarrow$ Volume / Speed $\rightarrow$} & \textbf{Observed} & \textbf{65 km/hr} & \textbf{55 km/hr} \\
    \hline
    \textbf{Observed} & V\_S  & V\_65 & V\_55 \\
    \hline
    \textbf{Half} & 1/2V\_S & 1/2V\_65 & 1/2V\_55 \\
    \hline
    \textbf{Double} & 2V\_S & 2V\_65 & 2V\_55 \\
    \hline
    \end{tabular}%
  \label{tab:scenarios}%
\end{table}%

\begin{figure}[!h]
\begin{subfigure}{\textwidth}
  \centering
  \includegraphics[width= 0.53\textwidth]{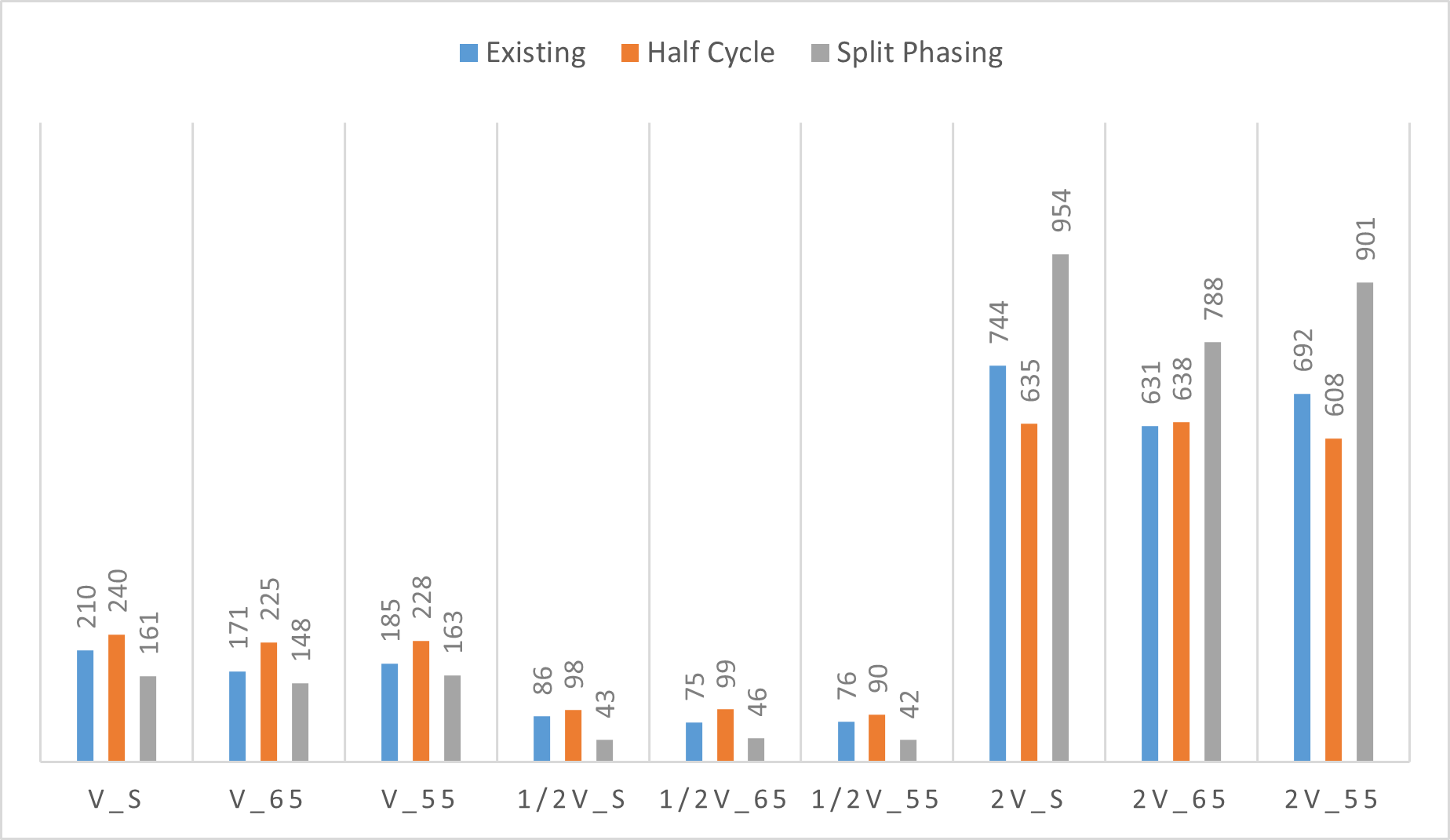}
  \caption{\label{fig:sub2} Count of car-following episodes.}
\end{subfigure}\\
\vspace{10pt}
\begin{subfigure}{\textwidth}
  \centering
  \includegraphics[width= 0.53\textwidth]{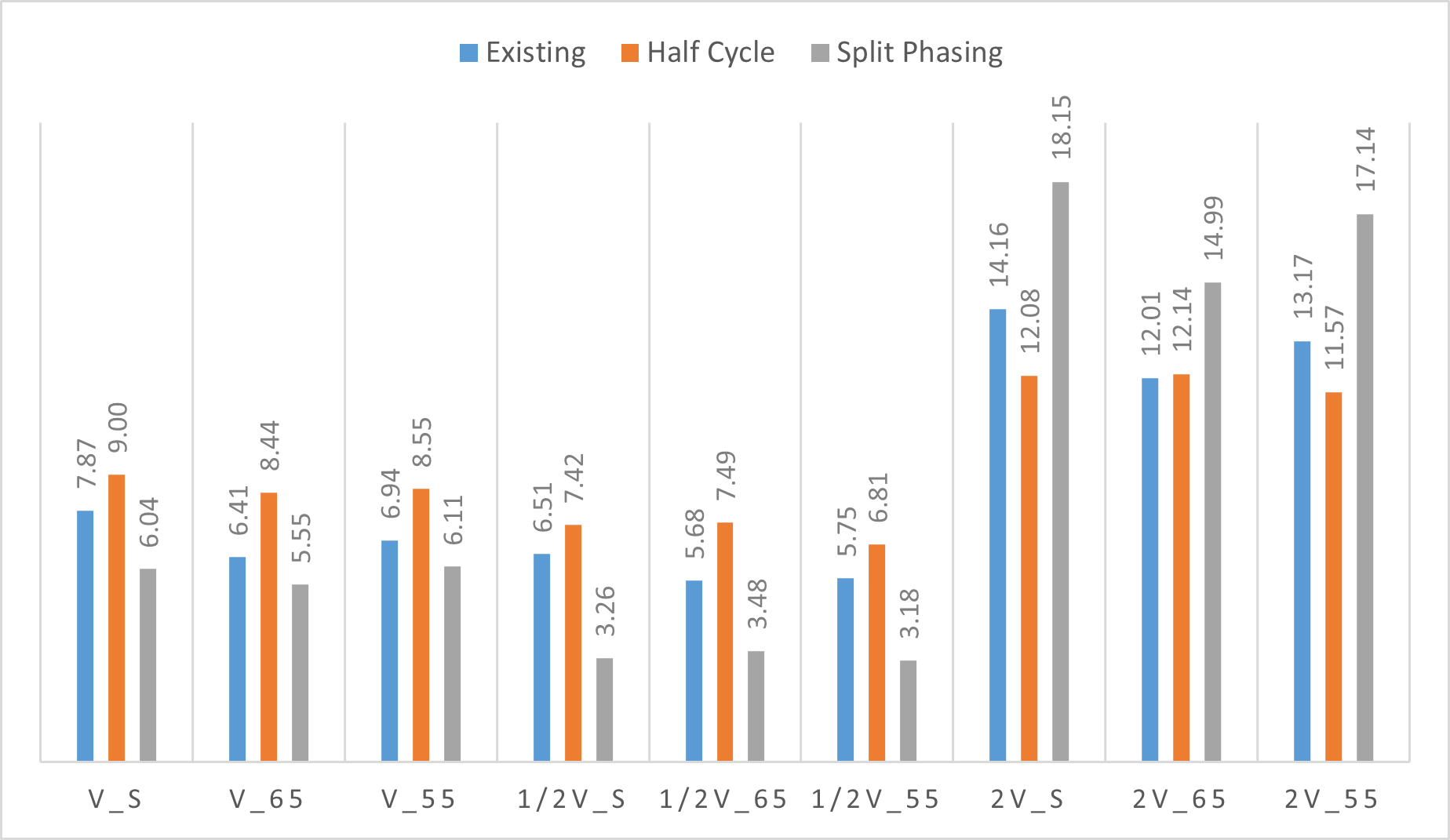}
  \caption{\label{fig:sub1} Percentage of episodes relative to total volume.}
\end{subfigure}
\caption{\label{fig:scenarios} Car-following episodes with safety-critical instantaneous TTC ($\leq$ 1.5 sec) for the different simulated scenarios.}
\end{figure}

Figure \ref{fig:scenarios} shows the counts and percentages of car-following episodes with instances of safety-critical TTC $\leq$ 1.5 sec. For the existing volume and given turn movement classifications at the intersection, utilizing a split-phasing control was found to reduce the number of generated safety-critical conflicts by over 23\%. Enforcing a 65 km/hr speed provides further but not substantial improvements. Split-phasing was also found to produce substantially better results in terms of safety in the scenario of lower traffic volumes. Conversely, it was the lowest performer in the case of double the observed volume.

\begin{figure}[!h]
\centering
\includegraphics[width= \textwidth]{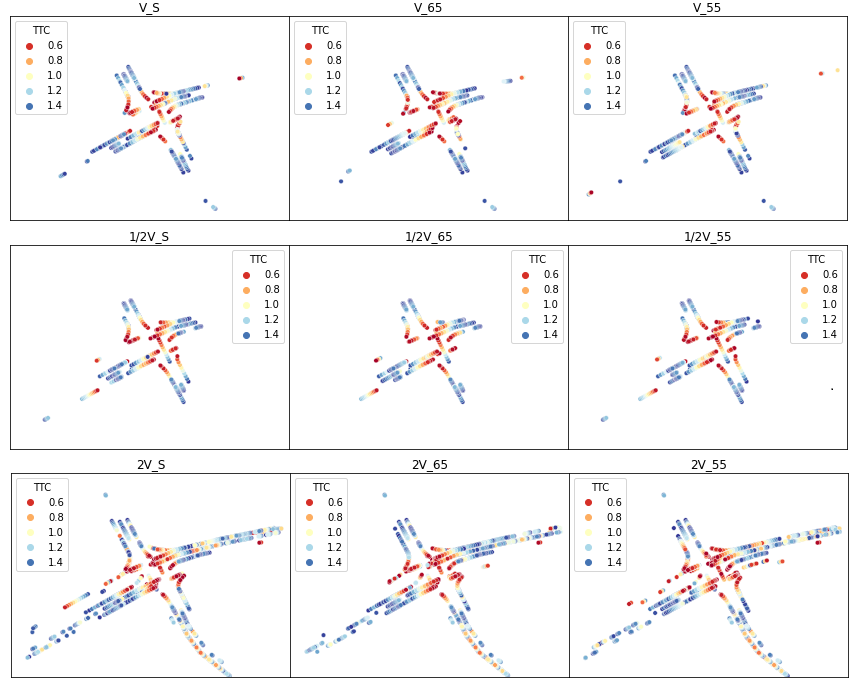}
\caption{\label{fig:heatmap1} Instantaneous safety-critical conflicts under current signal control.}
\end{figure}

\begin{figure}[!h]
\centering
\includegraphics[width= 0.74\textwidth]{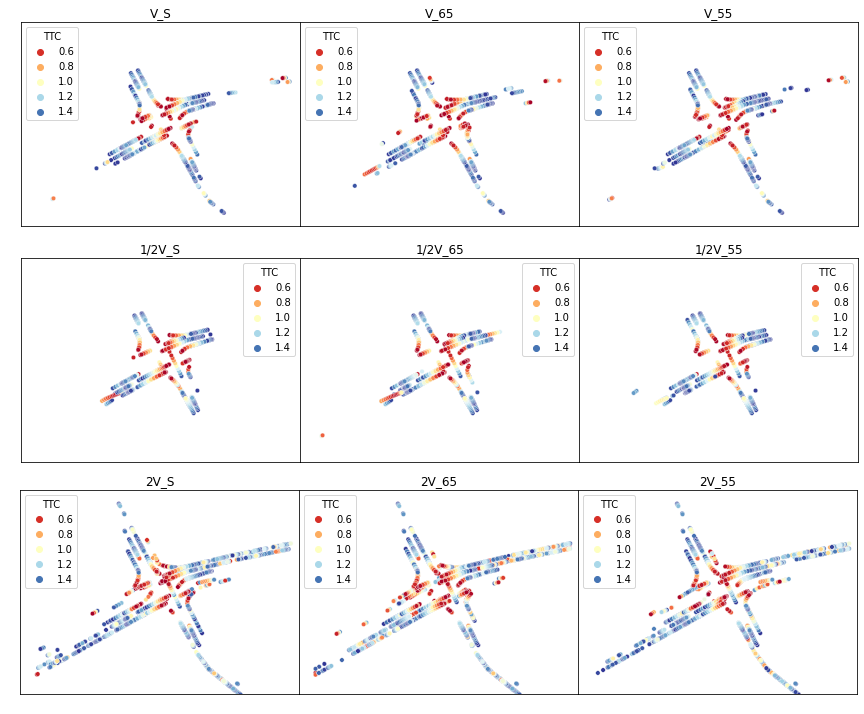}
\caption{\label{fig:heatmap2} Instantaneous safety-critical conflicts under half cycle length.}
\end{figure}

\begin{figure}[!h]
\centering
\includegraphics[width= 0.74\textwidth]{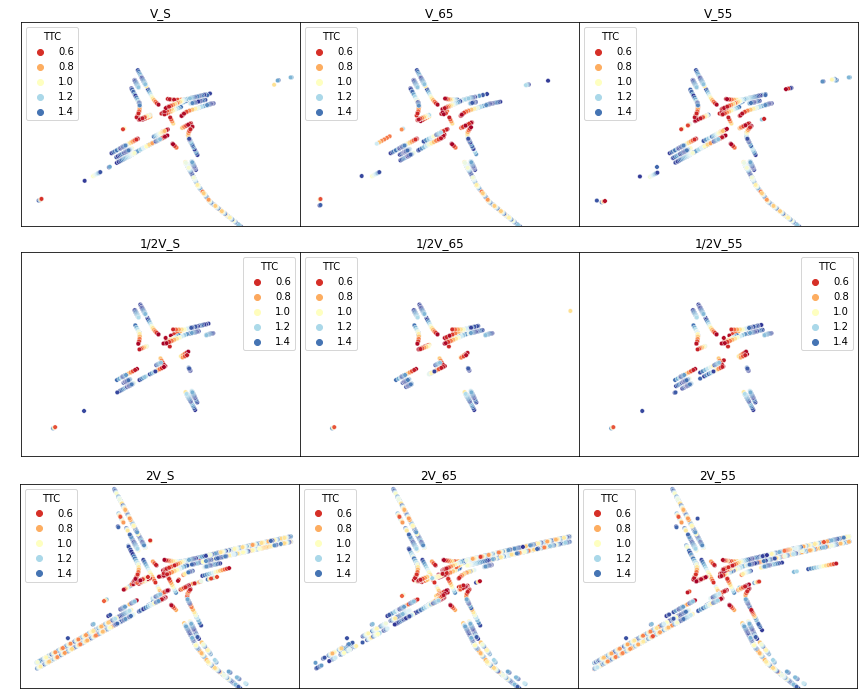}
\caption{\label{fig:heatmap3} Instantaneous safety-critical conflicts under split-phasing.}
\end{figure}

Figures \ref{fig:heatmap1} to \ref{fig:heatmap3} show all of those safety-critical conflicts for the different simulation scenarios, illustrating instantaneous (not averaged) safety-critical time-to-collision conflicts and their locations within the intersection's area. The split-phasing which was found to significantly reduce the existing safety conflicts can be seen in Figure \ref{fig:heatmap3} almost eliminating the conflicts generated at the turning lanes. The extended queues for phases 2 and 6 due to split-phasing resulted in an extended shockwave as vehicles approach and stop at the intersection during a red light, hence it is to be avoided in case of an increased traffic volume at this intersection.

\section{CONCLUSIONS}
This study provided an assessment of utilizing a data-driven driver behavior model to provide a high-fidelity simulation of safety-critical behavior. A safety-based Optimal Velocity Model that was calibrated for the location of study in Blacksburg, VA, and implemented in VISSIM as an external driver model. The resulting safety conflicts were compared to that of VISSIM's default Wiedemann model, and the results showcase the substantial improvements that can be achieved in simulation utilizing our proposed model in terms of replicating the existing safety concerns. We utilized this improved behavior modeling to quantify the impact of different signal control and speed enforcement scenarios in mitigating the simulated safety concerns. 

Unlike existing studies in the literature, our assessment did not involve a process of specifically calibrating VISSIM's model parameters to replicate the observed or expected safety concerns in the area of study. Conversely, our safety-based OVM utilized as an external driver model in this study was calibrated beforehand from video inference, and the direct implementation of that calibrated model resulted in a highly accurate representation of the status-quo at the area of study without targeted fine-tuning within VISSIM. The results highlight the value of this approach of video inference-based driver behavior modeling which provides a highly scalable, automated, and generalizable workflow and succeeds in better replicating the observed behavior in microsimulation. Vehicle trajectories can be efficiently extracted in real-time via our VT-Lane framework. While additional time is required post trajectory acquisition to calibrate the model parameters, the task can be effectively accomplished in near real-time which allows for the assessment of changes in driving behavior due to traffic disruptions, changes in weather, and other factors.

We concluded this study by assessing the potential impact of different signal control and speed enforcement measures to reduce the simulated conflicts at the intersection of the study across varying traffic volumes. The results obtained and discussed in this study do not take into account the trade-off between capacity and operational safety, which could be the subject of future works. Those results, however, do present the practitioners with an additional decision-support layer that effectively quantifies the expected safety impact for the selected interventions as a function of the existing driving behavior in an area of interest. Future studies will also assess the scalability and transferability of this approach, which could provide traffic safety practitioners and transport agencies with a robust decision-support tool for evaluating the safety impact of different traffic demand management, signal control, and enforcement measures.

\section{Author Contribution Statement}
The authors confirm their contribution to the paper as follows: study conception and design: A.A., M.A.; data collection: A.A.; analysis and interpretation of results: A.A.; draft manuscript preparation: A.A., M.A. Both authors reviewed the results and approved the final version of the manuscript. The authors do not have any conflicts of interest to declare.

\bibliographystyle{trb}
\bibliography{references}

\end{document}